%% file: main.tex
\documentclass{article}
\usepackage{iclr2027_conference,times}
\usepackage{inconsolata}
\input{math_commands.tex}

\usepackage{graphicx}
\usepackage{wrapfig}
\usepackage{capt-of}
\usepackage{placeins}
\usepackage{booktabs}
\usepackage{hyperref}
\usepackage{url}

\input{macros.tex}
\title{Late Attention Layers Alone Can Copy Entity Tokens, but Not Without Attending to Their Context}
\newcommand{\drexel}{\raisebox{-0.15ex}{\includegraphics[height=0.8em]{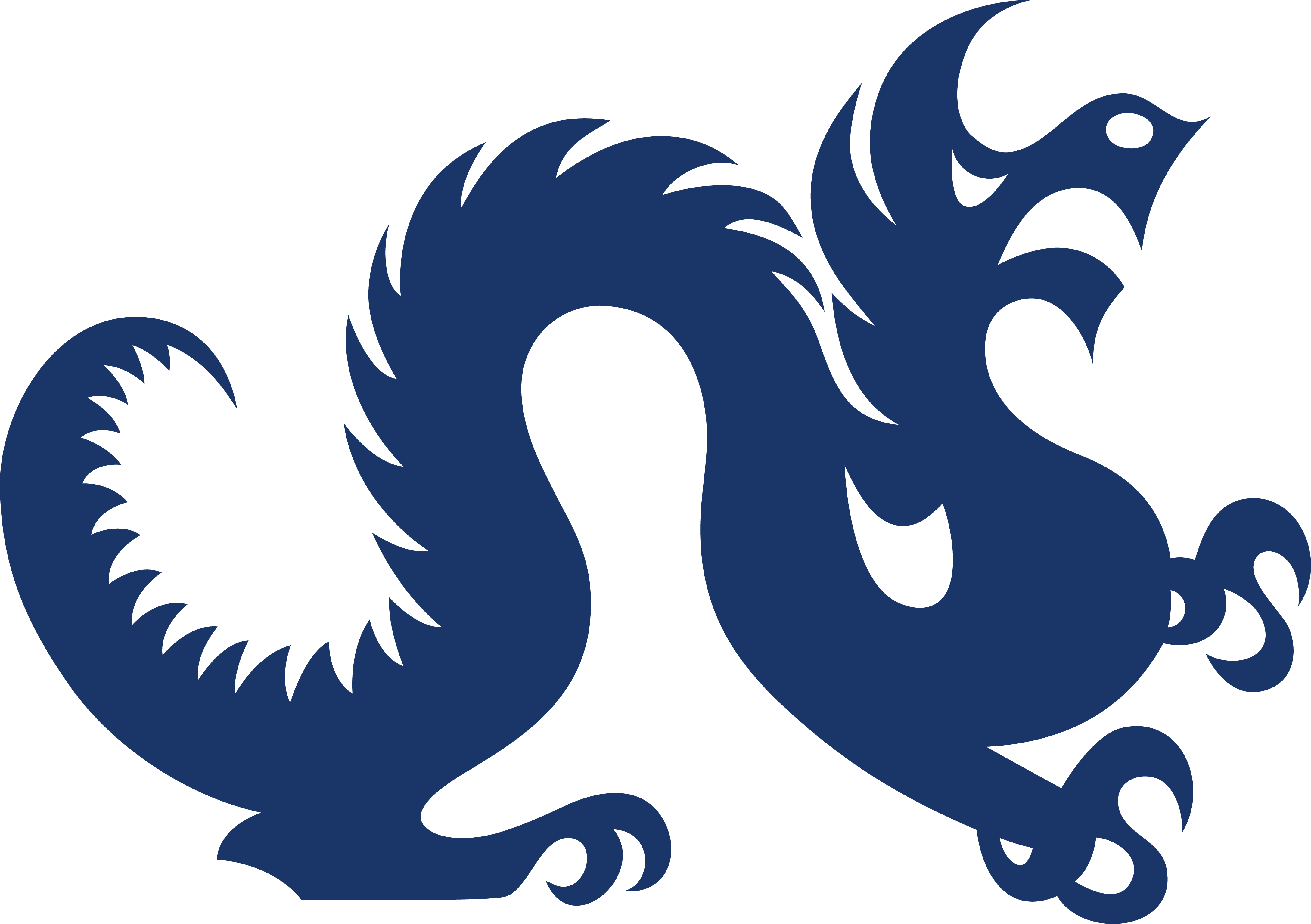}}}
\author{
Muyu He\hspace{0.4em}\drexel$^{*}$ \quad Yuchen Liu\hspace{0.4em}\drexel$^{*}$ \quad Ran Tao\hspace{0.4em}\drexel$^{*}$ \quad Li Zhang\hspace{0.4em}\drexel\\[0.4em]
\normalfont\normalsize \drexel\ Drexel University \qquad $^{*}$Equal contribution.
}
\makeatletter
\renewcommand{\@maketitle}{%
  \vbox{\hsize\textwidth
    \centering
    {\LARGE\bfseries\@title\par}
    \vskip 1.5em
    {\large\bfseries\begin{tabular}{@{}c@{}}\@author\end{tabular}\par}
    \vskip 0.3in minus 0.1in
  }%
}
\makeatother
\iclrfinalcopy
\hypersetup{
    pdftitle={Late Attention Layers Alone Can Copy Entity Tokens, but Not Without Attending to Their Context},
    pdfauthor={Muyu He, Yuchen Liu, Ran Tao, Li Zhang},
    hidelinks
}

\begin{document}
\maketitle
\fancyhead{}
\renewcommand{\headrulewidth}{0pt}
\begin{abstract}
\input{sections/00_abstract}
\end{abstract}
\input{sections/01_introduction}
\input{sections/02_setup}
\input{sections/03_attention_layers}
\input{sections/04_context_attention}
\input{sections/05_context_storage}
\input{sections/06_related_work}
\input{sections/07_conclusion}
\bibliography{references}
\bibliographystyle{iclr2027_conference}
\clearpage
\appendix
\input{sections/09_appendix}
\end{document}

%% file: math_commands.tex
\usepackage{amsmath,amsfonts,bm}

\def\eqref#1{equation~\ref{#1}}

\def\1{\bm{1}}

\DeclareMathAlphabet{\mathsfit}{\encodingdefault}{\sfdefault}{m}{sl}
\SetMathAlphabet{\mathsfit}{bold}{\encodingdefault}{\sfdefault}{bx}{n}

%% file: sections/00_abstract.tex
Large language models (LLMs) reliably perform \textit{entity copying},
in which a model copies tokens referring to an entity, termed \textit{entity tokens},
from the prompt into its output to answer a question.
Although entity copying is straightforward for most LLMs,
existing research does not provide a systematic account of
which layers specialize in this fundamental task
or how other tokens in the same sequence, termed \textit{context tokens},
influence the model's ability to copy the entity tokens.
To address these questions, we conduct experiments on Qwen3-8B
using two novel methods:
\textit{genie-in-a-bottle},
which controls exactly which layers can participate in an entity-copying task,
and \textit{attention lobotomy},
which cuts off specific tokens' attention to entity tokens
without affecting the remaining attention distribution.
We find that two distinct groups of layers in the second half of the model
are both necessary and sufficient
for entity copying.
Moreover, in addition to the decoding position's attention
to entity tokens,
context tokens' attention to entity tokens
also proves necessary for copying the exact tokens,
even though context tokens do not 
store entity information themselves
unless they satisfy particular semantic properties.
Our findings establish the critical role of late layers
in entity copying under the guidance of context tokens,
calling for future work on how models propagate and consume entity information.

%% file: sections/01_introduction.tex
\section{Introduction}
\label{sec:introduction}

Large language models (LLMs) built on the transformer backbone \citep{vaswani2017attention}
achieve their impressive performance
on natural language generation tasks
primarily through the attention mechanism,
which allows the current token position to retrieve information from earlier tokens in the sequence.
A particularly fundamental task that many models can perform is \textit{entity copying}
\citep{wang2023ioi,merullo2024reuse,tigges2024consistent,feucht2025dualroute,elhelo2025functionality},
in which the model needs to copy the exact tokens that refer to an entity,
or \textit{entity tokens},
from the prompt to the output
to answer a question about the entity correctly.
For example, the model needs to copy ``Einstein'' from the prompt
``Q: My friend is Einstein. What is my friend's name? A:''
to produce the output ``Einstein''.
Studying entity copying is critical
because it draws on the model's basic capacity
to extract the literal meaning of a token
and transmit it to a later token position,
an ability required for any task that involves retrieving information from context.
Although existing work has examined attention heads sensitive to entity copying \citep{elhelo2025functionality}
and the differences between copying concepts and copying literal words \citep{feucht2025dualroute},
insufficient attention has been paid
either to the role of groups of layers at different depths
or to that of tokens in the broader context that attend to entity tokens,
which we call \textit{context tokens}.

Recent work suggests that, outside the entity copying task,
different layers may specialize in entirely different subtasks \citep{meng2022locating,geva2023dissecting}.
For example, in tasks that require different layers to collaborate,
such as deciding when to insert a line break under a character-count constraint,
models have been found to use earlier attention layers
to determine the current character's position in the line
and later layers to calculate the distance from the character limit,
indicating layer specialization for each subtask \citep{gurnee2025when}.
Similarly, attention to context tokens has been found to contribute meaningfully
even when the context tokens themselves do not contain information to be retrieved.
For example, the mere presence of extra whitespace tokens can significantly increase
the model's ability to recall a particular fact to answer a question \citep{gekhman2026thinking},
and the verbalization of objects in an image
can help models perceive images more accurately \citep{lu2026visualprimitives}.
For the task of entity copying,
however, there is still insufficient work on
the role of specific layer groups and the impact of specific context tokens.

Existing methodologies are ill suited to
studying the roles of layers and context tokens in entity copying.
The standard method for attributing roles to layers in entity information retrieval,
which injects hidden states from a layer attending to the entity into an uncontaminated prompt
\citep{ghandeharioun2024patchscopes,cheng2026engram},
falls short of controlling for confounds such as the number of layers available
to complete the forward pass after the injection.
As a result, we propose our first method, \textit{genie-in-a-bottle},
to capture the transmission of entity information
within a fixed sliding window of activated layers,
ensuring that each layer's contribution is carefully controlled.
Moreover, existing techniques study context tokens' influence naively
by enabling or disabling the model's attention to them \citep{geva2023dissecting}.
As such, they cannot disentangle context tokens' role in supplying
relevant semantic information themselves
from their more implicit contributions, such as influencing how strongly the model attends to other tokens.
As a remedy, we propose our second method,
\textit{attention lobotomy},
to sever selected tokens' attention to the entity token
while preserving their complete hidden states,
thereby providing a clean way to measure the causal effect of the content of their states
on the model's entity copying ability.

Applying the two techniques to Qwen3-8B \citep{qwen2025qwen3}, we uncover causal patterns
in its entity copying computations
associated with specific groups of layers and context tokens.
Regarding layer roles, we find that entity copying is handled almost entirely
by two distinct groups of layers in the second half,
with minimal contributions from layers in the first half
and a region near the end.
Regarding context tokens,
we find that their attention to entity tokens is often required
in addition to the decoding position's own attention to those tokens.
Without context tokens' attention to the entity,
the model often fails to recover the exact tokens referring to it
and is sometimes unable to extract any entity information at all.
In addition, we find that although context tokens can influence entity-copying success,
they do not themselves store entity information
unless they have very specific semantic properties:
context tokens that are factually correct about the entity
(for example, ``a physicist'' rather than ``an athlete'' for ``Einstein'')
and closely associated with it
(for example, ``a physicist'' rather than ``a human'')
can sometimes store enough entity information to support copying.
Together, these findings show that entity copying
activates a highly specialized group of late layers
that awaits directions from context tokens to allocate the attention budget.

We summarize our key findings below:
\begin{itemize}
\item \textbf{Two interpretability methodologies to control causal factors in attention computation.}
We propose \textit{attention lobotomy} and \textit{genie-in-a-bottle}
to study the contributions of isolated features,
such as attention from context tokens or groups of layers,
to entity copying tasks.
\item \textbf{The specialized role of late layers in entity copying.}
We establish that two groups of layers in the second half of Qwen3-8B
are necessary and sufficient for entity copying.
\item \textbf{The guiding role of context tokens in determining the strength of attention to the entity.}
We find that context tokens aid entity copying
not by storing entity information in their hidden states
but by guiding the model through their joint attention to the entity token.
They can store entity information only if they have certain semantic properties.
\end{itemize}

%% file: sections/02_setup.tex
\section{Setup}
\label{sec:setup}

\begin{figure}[t]
  \centering
  \includegraphics[width=\linewidth]{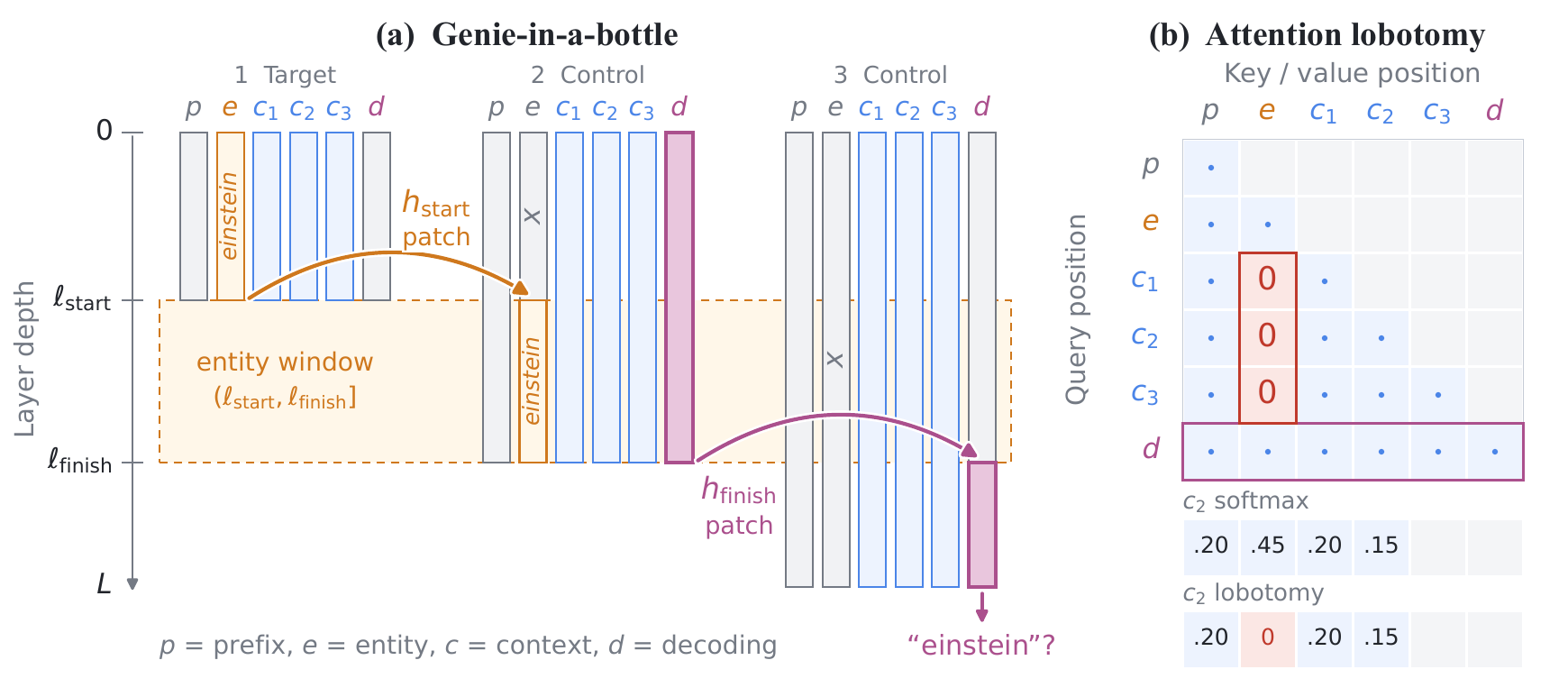}
  \caption{\textbf{Two methodologies for studying entity copying.}
  \textbf{(a)} Genie-in-a-bottle confines access to the entity token
  $E$ to the layer window from $\ell_{\mathrm{start}}$ to $\ell_{\mathrm{finish}}$
  by performing three separate forward passes
  before, within, and after the window.
  Only the second pass, within the window, enables attention to $E$.
  \textbf{(b)} Attention lobotomy removes specific attention pathways
  $x \to y$ to study their effect on entity copying.
  The effective attention weights for selected destination tokens $y$ are zeroed after softmax.}
  \label{fig:methods}
\end{figure}

\begin{figure}[!htb]
    \centering
    \includegraphics[width=0.94\linewidth]{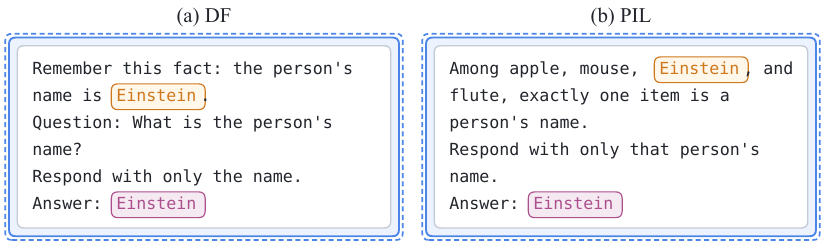}
    \caption{Two of the five prompt templates, shown with the same entity and expected answer.}
    \label{fig:template-examples}
\end{figure}

\paragraph{Datasets.}
To study how the model copies diverse entity tokens
across different contexts and cognitive tasks,
we collect 100 entities and place them in five prompt templates.
We partition the 100 entities into two groups:
50 consist of a single token, such as ``Messi'' or ``Einstein'',
and the other 50 consist of two tokens, such as ``Emily Dickinson''.
We study both groups to control for any changes in
the model's strategy
arising from its need to copy more than one token from the context.
All entities are placed at the same location within a given template
(\texttt{direct\_fact} (DF), \texttt{visitor\_register} (VR), \texttt{friend} (F), \texttt{name\_badge} (NB), or \texttt{person\_in\_list} (PIL)),
each of which poses a natural-language question whose answer depends on copying the entity tokens (Figure~\ref{fig:template-examples}).
Appendix~\ref{app:inputs} provides the complete templates and entity lists.

\paragraph{Model.}
All experiments use Qwen3-8B-Base \citep{qwen2025qwen3}
in completion mode without a chat template
to frame entity copying as a next-token prediction task.
The model has 36 transformer blocks,
each consisting of a full grouped-query attention (GQA) layer \citep{ainslie2023gqa}
and an MLP layer.
Each attention layer has 32 query heads and 8 key/value heads.
We use greedy decoding (temperature zero) for reproducible completions,
as stochasticity is irrelevant to this study.

\paragraph{Metrics.}
We judge the success of entity copying for each entity case
by whether the correct entity tokens occur in the model's completion
of up to 12 tokens.
We classify outputs into five categories:
\texttt{exact\_name} copies the exact entity tokens (such as ``einstein'' for ``einstein'');
\texttt{nonexact\_name} recovers tokens that do not exactly match the entity tokens
but nevertheless correctly refer to the entity (``albert'' for ``einstein'');
\texttt{wrong\_name} produces tokens that refer to another entity (``newton'' for ``einstein'');
\texttt{not\_name} outputs tokens that are coherent but do not include any entity tokens (``i do not know'');
and \texttt{malformed} produces tokens that are ungrammatical,
signaling a disruptive intervention during output generation.
For each template, we report both
\texttt{exact\_success\_rate}, the rate at which the model produces \texttt{exact\_name}, and
\texttt{general\_success\_rate}, the rate at which the model produces either \texttt{exact\_name} or \texttt{nonexact\_name},
and monitor abnormal proportions of \texttt{malformed} outputs.

\paragraph{Methodologies.}
We introduce two interpretability techniques, illustrated in Figure~\ref{fig:methods},
to study the roles of layer groups and context tokens in entity copying.

We propose \textit{genie-in-a-bottle} to determine
whether attention layers within a fixed window from $\ell_{\mathrm{start}}$ to $\ell_{\mathrm{finish}}$
directly extract entity information
and, if not, whether these layers contribute indirectly to entity copying
through some other mechanism.
For a given entity to copy, we prepare two sequences:
a target sequence in which the entity tokens are present
(``Q: einstein is my friend. who is my friend. A:'')
and a control sequence in which the entity tokens are replaced by a control token
(``Q: x is my friend. who is my friend. A:'').
Then, to confine access to the entity tokens
to the window from $\ell_{\mathrm{start}}$ to $\ell_{\mathrm{finish}}$,
we run three forward passes sequentially:
first, we run the target sequence up to $\ell_{\mathrm{start}}$
and extract the hidden state $h_{\mathrm{start}}$ at the entity token position;
second, we patch $h_{\mathrm{start}}$ into the corresponding token position at $\ell_{\mathrm{start}}$ in the control sequence,
run the control sequence up to $\ell_{\mathrm{finish}}$,
and extract the hidden state $h_{\mathrm{finish}}$ at the decoding token position;
third, in a fresh forward pass of the control sequence, we patch $h_{\mathrm{finish}}$ into the same token position at $\ell_{\mathrm{finish}}$
and continue the forward pass and decode the output
to assess copying success.
Across these three forward passes,
the last token position can extract entity information only
within the window from $\ell_{\mathrm{start}}$ to $\ell_{\mathrm{finish}}$.
Furthermore, to prevent entity information from
leaking into context tokens outside the window,
context tokens cannot attend to the entity token position in
either the target or the control sequence.

We also propose \textit{attention lobotomy} to determine
whether a particular attention pathway
between two tokens in the context
causally affects entity copying.
In particular, we examine whether attention from context tokens
to entity tokens matters,
and whether the decoding token position needs to attend to context tokens at all.
To determine the impact of a given attention pathway $x \to y$
from token $x$ to token $y$,
we set the effective attention weight for $y$ to zero
after computing the softmax attention weights from $x$ to all previous tokens.
We then inspect the decoded outputs to see whether entity copying is still successful.
Zeroing the effective attention weight for $y$ after softmax
ensures that the attention weights for all other tokens $t \ne y$ remain unchanged at that operation.
Otherwise, any degradation in the decoded outputs can be attributed to
the confounding effect of inappropriate attention to tokens $t \ne y$.

%% file: sections/03_attention_layers.tex
\section{Which layers are critical for entity copying?}
\label{sec:attention-layers}

We study the role of layers at different depths in entity copying.
In particular, we ask for which windows of consecutive layers attention to entity tokens
is necessary, meaning that removing it would substantially reduce entity-copying success,
or sufficient, meaning that retaining it alone, without support from other layers,
is enough to substantially improve copying success.
In addition, for layers that prove neither necessary nor sufficient,
we explore whether computations other than their direct attention to entity tokens
are still necessary for entity copying to succeed.

\begin{figure}[!htb]
    \centering
    \includegraphics[width=\linewidth]{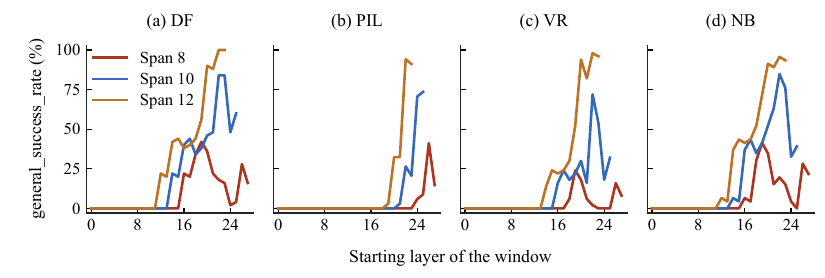}
    \caption{
        The \texttt{general\_success\_rate} when only a fixed window of
        attention layers can attend to entity tokens,
        for different window spans.
    }
    \label{fig:relay-blocked}
\end{figure}

\paragraph{Layers in the second half are sufficient for extracting entity information, with two isolated groups contributing the most.}
We apply genie-in-a-bottle to four of the five templates
after filtering (see Appendix~\ref{app:genie-filtering}),
using fixed window sizes of 8, 10, and 12 layers (Figure~\ref{fig:relay-blocked}).
To ensure that entity extraction comes directly
from attention to the entity tokens within the fixed window of layers,
we remove the decoding position's attention to the entity position
outside the window, where that position contains the control token
rather than the target entity token by design.

As shown in Figure~\ref{fig:relay-blocked}, we draw several conclusions
about how groups of layers contribute to entity extraction.
(1) Entity extraction occurs in the second half of the layers.
For window sizes of 8, 10, and 12,
windows in the first half yield a \texttt{general\_success\_rate} of zero.
By contrast, windows in the second half
produce correct answers, indicating successful entity extraction.
As the window size increases,
the maximum \texttt{general\_success\_rate} for that window size
approaches full recovery,
suggesting that layers in the second half are responsible for entity extraction.
(2) A mid-late group of layers contributes significantly to entity extraction.
Around L20, \texttt{general\_success\_rate} peaks for a window size of 8,
while it begins to rise much more steeply for window sizes of 10 and 12.
This increase in successful outputs indicates that
a group of layers around L20 is critical for entity extraction.
(3) A late group of layers contributes minimally.
Around L24, \texttt{general\_success\_rate} plummets
for window sizes of 8 and 10 before recovering modestly,
suggesting that a group near the end plays only a minor role in entity extraction on its own.
(4) A second group at the end also contributes significantly.
We find that \texttt{general\_success\_rate} reaches its maximum near the end
for a window size of 12, while it recovers from an earlier collapse
for window sizes of 8 and 10.
Since a window of size 12 in this region cannot rely on the earlier region around L20,
its improved performance results from a second group
that contributes substantially to entity extraction.

\begin{figure}[!htb]
    \centering
    \includegraphics[width=\linewidth]{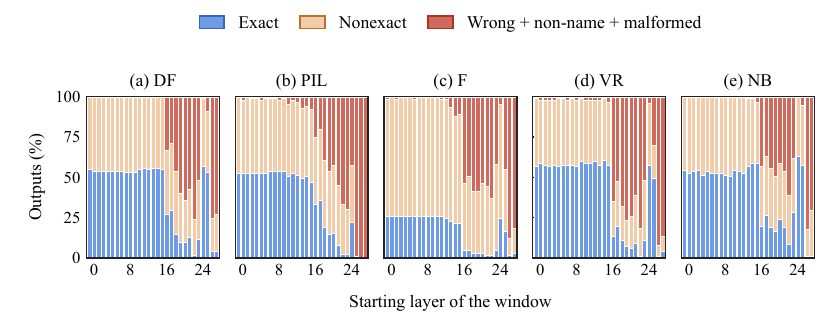}
    \caption{
        Output distribution after removing attention to entities
        within a fixed window.
    }
    \label{fig:attention-necessity-width8}
\end{figure}

\paragraph{The same two discrete groups in the second half are also necessary for entity extraction.}
We apply attention lobotomy to the five templates using a window size of 8,
removing attention to the entity tokens within each window
(Figure~\ref{fig:attention-necessity-width8}).
We find that removing attention within windows in the first half of the layers
produces few or no \texttt{wrong\_name}, \texttt{non\_name}, or \texttt{malformed} outputs.
By contrast, removing attention within windows in the second half
causes significant degradation.
This establishes that most layers in the second half are also necessary
for entity extraction, in addition to being generally sufficient.
Moreover, we observe the highest failure rates when removing attention
in two distinct groups of layers: one just past the start of the second half
and one at the end.
By contrast, removing attention in a group of layers near L24
causes almost no failures in four templates
and relatively few in the remaining template compared with neighboring windows.
This corroborates the preceding finding that two discrete groups of layers
in the second half, separated by a region that contributes less,
are not only sufficient but also necessary for entity copying in most cases.

\begin{figure}[!htb]
    \centering
    \includegraphics[width=\linewidth]{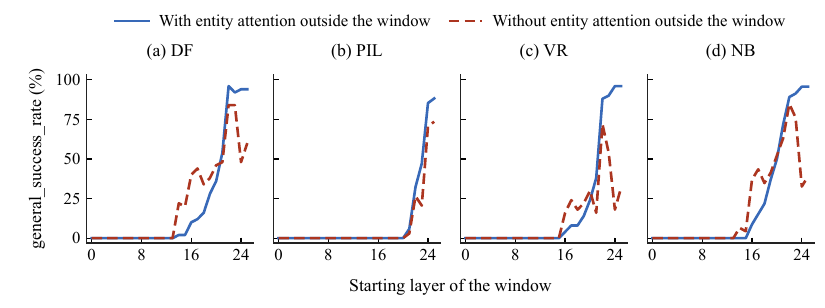}
    \caption{
        Comparison of \texttt{general\_success\_rate}
        with attention to the control entity enabled versus disabled
        outside the designated window.
    }
    \label{fig:relay-span10-cohorts}
\end{figure}

\paragraph{Although early layers do not participate in entity extraction, their attention to the same position is critical for deep layers to extract entity information successfully.}
We run genie-in-a-bottle again on the same four templates
with a window size of 10,
but instead of removing attention to the entity position outside the window,
we allow layers to attend to that position,
which contains the confounding control token.
As a result, layers outside the window attend to the control token
and write their outputs back into the residual stream,
where they compete with information about the entity token.
Although we expect degradation across all windows
because of this shared confound,
we find that \texttt{general\_success\_rate} generally increases across windows (Figure~\ref{fig:relay-span10-cohorts}).
Windows deeper in the second half benefit particularly
from early layers attending to the entity position containing the competing token,
with higher success rates and no signs of collapse as the windows move deeper.
Since the control token is randomly selected for each run
and bears no semantic similarity to the target entity token
(e.g., ``Einstein'' versus ``Messi''),
this suggests that early layers' attention to the same token position
is critical to later layers' ability to extract entity information.

%% file: sections/04_context_attention.tex
\section{Can the model copy the exact entity tokens without context tokens attending to them?}
\label{sec:context-attention}

\begin{figure}[htbp]
    \centering
    \includegraphics[width=\linewidth]{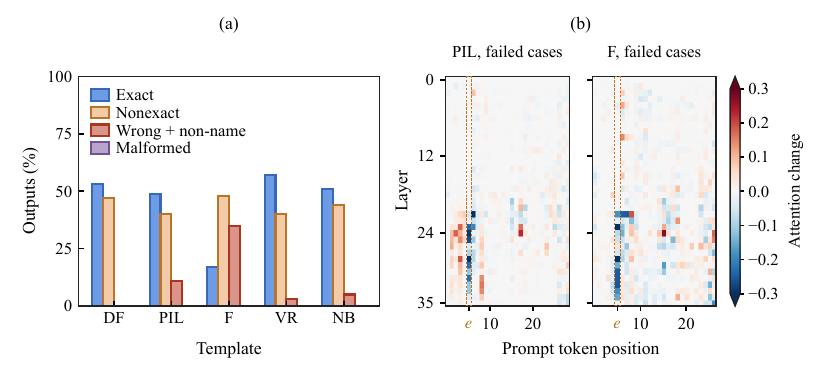}
    \caption{
        (a) Breakdown of output categories for each template
        after removing context tokens' attention to entities.
        (b) For the two templates with the most unsuccessful outputs,
        the decoding position's attention to entities decreases sharply
        after context tokens stop attending to them.
    }
    \label{fig:context-copying}
\end{figure}

\paragraph{The model needs context tokens' joint attention to the entity to copy the literal entity tokens, but it can recover the concept of the entity by attending to the entity alone.}
We perform attention lobotomy by removing the context tokens' attention
to the entity token while retaining every other attention pathway.
In particular, the decoding position can still attend to
both the context tokens to determine the required output format
and the entity tokens to copy them one by one,
but the context tokens' keys and values can no longer encode
any information about the entity.
We find that, despite being able to attend to each individual entity token,
the decoding position has trouble copying the exact tokens in the prompt (Figure~\ref{fig:context-copying}(a)).
Instead, almost half the time, the model outputs
\texttt{nonexact\_name} responses, such as
``George'' for ``Orwell'' (F) and
``J.\ R.\ R.\ Tolkien'' for ``Tolkien'' (NB).
Thus, without context tokens' joint attention to the entity,
the model often outputs a different set of tokens referring to the same entity,
extracting only the concept of the entity rather than its exact tokens.
This indicates that context tokens' attention to the same entity
is often necessary for copying the literal entity tokens,
and that copying entity tokens and extracting the concept of the entity
require different computations.

\paragraph{Removing context tokens' attention to the entity can also reduce subsequent positions' attention to the same entity.}
For F and PIL, two of the five templates to which we apply attention lobotomy,
we find a particularly high proportion of \texttt{wrong\_name} and
\texttt{non\_name} outputs,
whereas \texttt{exact\_name} and \texttt{nonexact\_name} outputs
predominate in the other three templates (Figure~\ref{fig:context-copying}(a)).
For example, when context tokens cannot attend to ``Nobel'' in PIL,
the model incorrectly answers ``Mouse'' when asked to identify the person's name.
This suggests a different causal mechanism underlying the unsuccessful outputs.
At each layer, for each context token attended to by the decoding position,
we plot the change in the maximum attention score across heads, averaged across examples,
in Figure~\ref{fig:context-copying}(b), showing how peak attention at each layer changes
after attention lobotomy on context tokens.
We find a large reduction in the late layers' attention to the entity tokens
for both templates, in contrast to little or no change in attention
for the other three templates.
When we restore the decoding position's attention to the entity in each head
to its value before attention lobotomy,
almost all outputs return to \texttt{exact\_name} or \texttt{nonexact\_name}
rather than the unsuccessful categories.
This indicates a causal chain for the two templates:
removing context tokens' attention to the entity tokens
can directly reduce subsequent positions' attention to the same entity,
and this reduction in attention in turn produces unsuccessful outputs.

\paragraph{Context tokens need only 1--2 layers of attention to deposit information critical for successful entity copying.}
We investigate which layers enable context tokens to influence
entity-copying success.
We therefore select three groups of layers at different depths:
an early group (L5--6), a mid group (L12--13), and a late-mid group (L19--20).
Following the previous attention lobotomy on context tokens in F and PIL,
we restore attention from context tokens to entity tokens
in only one group at a time (Figure~\ref{fig:context-restoration-storage}(a)).
We find that, on both templates, restoring attention in any of the three groups
substantially improves the model's \texttt{exact\_success\_rate}.
In particular, the early group (L5--6) appears to be the most effective,
as it substantially improves performance on both templates.
This suggests that, at different layer depths,
context tokens' attention to entity tokens deposits information
critical for entity copying in their key and value vectors.

%% file: sections/05_context_storage.tex
\section{Do context tokens store entity information in their own states?}
\label{sec:context-storage}
Given our finding that context tokens' joint attention to entity tokens
causally affects subsequent tokens' ability to copy those tokens,
an intuitive hypothesis is that context tokens use this attention
to store entity information in their own states.
At decoding time, the decoding position can therefore retrieve some entity information
from context tokens, improving its copying success rate.
We design ablations to test this hypothesis against
the competing, less committal hypothesis that context tokens' attention to entities
merely guides subsequent attention and does not help store entity information in the context.

\begin{figure}[htbp]
    \centering
    \includegraphics[width=0.49\linewidth]{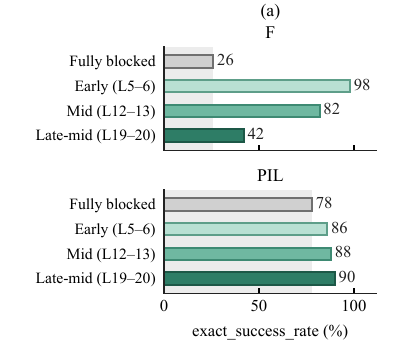}\hfill
    \includegraphics[width=0.49\linewidth]{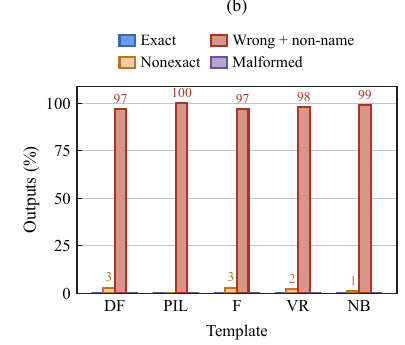}
    \caption{
        (a) Restoring context tokens' attention to entity tokens in just
        two layers improves subsequent positions' copying success.
        (b) With attention to entity tokens disabled
        and attention to context tokens intact, the model
        generates mostly wrong names or non-name responses.
    }
    \label{fig:context-restoration-storage}
\end{figure}

\paragraph{Despite their causal role in entity copying, context tokens themselves generally do not store entity information.}
To test the hypothesis that context tokens store entity information in their own states,
we apply attention lobotomy to the decoding position's attention to the entity tokens.
As a result, the decoding position can attend only to context tokens.
If context tokens store entity information in their own states,
the model should be able to retrieve that information from them
and achieve modest success rates in entity copying.
However, we find that almost all outputs under attention lobotomy
are \texttt{wrong\_name} or \texttt{non\_name}
(\mbox{Figure~\ref{fig:context-restoration-storage}(b)}).
This indicates that, despite contributing causally to successful entity copying
through their attention to entity tokens,
context tokens such as those in the five templates generally do not store entity information themselves.
Instead, they merely guide subsequent tokens to extract entity information effectively.

\begin{wrapfigure}{r}{0.44\linewidth}
    \centering
    \includegraphics[width=\linewidth]{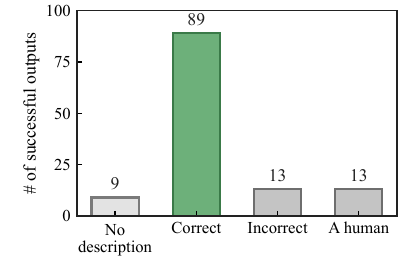}
    \caption{
        Number of successful outputs under four context conditions
        with attention to entity tokens removed.
    }
    \label{fig:description-recovery}
\end{wrapfigure}

\paragraph{Description tokens in the context can store entity information only if they are factually correct about the entity.}
Although context tokens generally do not store entity information,
we examine whether the semantic features of particular token types
can nonetheless predict their ability to do so.
Specifically, while removing the decoding position's attention to entities through attention lobotomy,
we append a short indefinite description of each entity's profession after the entity tokens,
such as ``a physicist'' for ``Einstein'' or ``a writer'' for ``Shakespeare'',
to study whether the description can store entity information.
An indefinite description does not uniquely identify an entity,
such as ``a physicist'' for either ``Einstein'' or ``Newton''.
By using an indefinite description in the prompt,
we ensure that successful copying of the entity tokens cannot result
from the model guessing the entity from the description.
We find that, although most outputs remain incorrect, as they do without the description,
the absolute number of successful outputs increases substantially
(\mbox{Figure~\ref{fig:description-recovery}}).
This indicates that, under specific circumstances,
indefinite descriptions do store entity information.

We then rotate the indefinite descriptions among entities,
so that each entity is paired with an incorrect description.
We find that copying performance does not improve relative to the condition without a description
and is significantly lower than with a correct description
(\mbox{Figure~\ref{fig:description-recovery}}).
This indicates that description tokens can store entity tokens
only if they state correct facts about the entity they describe.

\paragraph{Description tokens in the context can store entity information only if they are closely associated with the entity.}
As an alternative to appending indefinite descriptions of profession and social identity
after entity tokens, we append the fixed phrase ``a human''
after all entity tokens referring to people.
``A human'' is also factually correct for every entity it describes,
but differs from the preceding indefinite descriptions in being intentionally broad
and not closely associated with the entity in a practical sense.
We find minimal improvement in the number of successful outputs
relative to the condition without a description,
and performance is comparable to that with an incorrect description
(\mbox{Figure~\ref{fig:description-recovery}}).
This indicates that factual correctness is not a sufficient condition
for descriptions to store entity information.
Instead, it suggests that a close association with the entity
may be a prerequisite for a description to store entity information.

%% file: sections/06_related_work.tex
\section{Related Work}
\label{sec:related-work}

\paragraph{Copying mechanisms in language models.}
Mechanistic studies have identified attention heads that copy information from earlier
positions. Induction heads support sequence continuation by matching a repeated
prefix and promoting the token that previously followed it \citep{olsson2022induction}.
In indirect object identification, name-mover heads copy a name selected through
interactions with upstream heads \citep{wang2023ioi}.
Subsequent work finds reuse of these components across tasks
\citep{merullo2024reuse} and consistency of task-level mechanisms across training
and model scale \citep{tigges2024consistent}.
More recent studies distinguish token-level from concept-level copying
\citep{feucht2025dualroute} and infer copying and other head functions from
parameter mappings \citep{elhelo2025functionality}.
These results establish that copying involves both selecting a source and
transferring its content. We examine how these processes depend on layer groups
and context tokens, distinguishing recovery of an entity's identity from
reproduction of its exact name in the prompt.

\paragraph{Entity representations and contextual readout.}
Entity binding studies ask how models associate entities with attributes supplied
in context. \citet{feng2024binding} identify representations that support these
associations, while \citet{gurarieh2026mixing} distinguish positional, lexical,
and reflexive mechanisms for retrieving bound entities.
Recovering a name from an entity representation is a related but distinct problem:
\citet{morand2025representations} show that contextual information can improve
name reconstruction and distinguish exact mention recovery from responses that
refer to the same entity.
At the level of attention computation, \citet{kamath2025tracing} trace how
query- and key-side features jointly determine attention patterns.
Our experiments distinguish two possible roles of context tokens in entity copying:
providing information from which the answer can be recovered, and influencing
the final input position's attention to the entity itself.

\paragraph{Layer-wise retrieval of factual information.}
Studies of parametric knowledge also find distinct roles for entity representations
and later computations. \citet{meng2022locating} localize factual associations
using causal tracing, identifying important contributions from subject-position
MLPs and later attention layers.
\citet{geva2023dissecting} further separate subject enrichment, relation processing,
and attribute extraction during factual recall.
\citet{ferrando2025entity} show that intervening on representations of entity
familiarity changes subsequent attention and answering behavior.
These studies motivate examining when entity information becomes available to
the output position. Our setting differs in the information to be retrieved:
the requested name is already present in the prompt, rather than an attribute
that must be recalled from model parameters.

\paragraph{Causal interventions on attention and representations.}
Our interventions build on methods for measuring information flow and decoding
hidden states. Value Zeroing removes a token's value vector while retaining its
key and query, and measures the resulting change in contextual representations
\citep{mohebbi2023mixing}.
Patchscopes uses a language model to interpret representations transferred between
computations \citep{ghandeharioun2024patchscopes}; cross-prompt patching has also
been used to separate concept information from output language
\citep{dumas2025tongue}.
The interpretation of such interventions depends on the corruption, evaluation
metric, and patching window \citep{zhang2024patching}.
We combine query- and layer-specific, post-softmax removal of entity-value contributions with
a relay that transfers an entity state and then the resulting final-input-position
state at two layer boundaries. For window-only readout, same-window,
no-replacement controls identify cases
where restricted entity access already prevents identity recovery.
This setup tests whether an injected identity can affect the answer through a
specified window, without treating decodability alone as evidence of its use.

%% file: sections/07_conclusion.tex
\section{Conclusion}
\label{sec:conclusion}

We study the roles of layer groups at different depths
and of context tokens in Qwen3-8B's ability
to perform entity copying tasks.
We propose two interpretability techniques:
using genie-in-a-bottle,
we isolate a fixed-size window of layers
to assess each layer's contribution;
using attention lobotomy,
we remove individual attention pathways
to isolate the causal effects of context tokens' attention operations.
Regarding layers,
we find that two groups of layers in the second half are both
necessary and sufficient for extracting entity information,
whereas layers in the first half contribute indirectly by
consistently attending to the same token position.
Regarding context tokens,
we find that their attention to entity tokens
causally affects subsequent tokens' ability
to copy those same tokens,
even though context tokens do not
actually store entity information themselves
unless they have particular semantic relationships with the entity.

%% file: sections/09_appendix.tex
\section{Prompt templates, entities, and filtering}
\label{app:inputs}
\label{app:genie-filtering}

\subsection{Prompt templates}
Table~\ref{tab:templates} gives the five complete prompts.
We replace \texttt{\{entity\}} with a name from Table~\ref{tab:entities},
retain the line breaks shown, and generate directly after \texttt{Answer:}
without a chat template.

\begin{table}[!htb]
\centering
\caption{Complete prompt templates. Line breaks within each prompt are preserved.}
\label{tab:templates}
\small
\setlength{\tabcolsep}{6pt}
\begin{tabular}{@{}lp{0.88\linewidth}@{}}
\toprule
Template & Prompt \\
\midrule
DF & \ttfamily Remember this fact: the person's name is \{entity\}.\newline Question: What is the person's name?\newline Respond with only the name.\newline Answer: \\
\addlinespace[7pt]
PIL & \ttfamily Among apple, mouse, \{entity\}, and flute, exactly one item is a person's name.\newline Respond with only that person's name.\newline Answer: \\
\addlinespace[7pt]
F & \ttfamily Here is a fact: \{entity\} is my friend.\newline Question: What is my friend's name?\newline Respond with only the name.\newline Answer: \\
\addlinespace[7pt]
VR & \ttfamily The visitor signed the register with the name \{entity\}.\newline Question: What name did the visitor write?\newline Respond with only the name.\newline Answer: \\
\addlinespace[7pt]
NB & \ttfamily The name printed on the badge is \{entity\}.\newline Question: What name is printed on the badge?\newline Respond with only the name.\newline Answer: \\
\bottomrule
\end{tabular}
\end{table}

\clearpage
\subsection{Entity lists}
Table~\ref{tab:entities} lists all 100 entity strings with their original spelling
and capitalization.
The token counts refer to the Qwen3-8B-Base tokenizer in these prompts;
they are not word counts.
Experiments using genie-in-a-bottle use the one-token group before filtering.

\begin{table}[!htb]
\centering
\caption{Complete entity lists, grouped by entity-span length.}
\label{tab:entities}
\small
\setlength{\tabcolsep}{5pt}
\begin{tabular*}{\linewidth}{@{\extracolsep{\fill}}llll@{}}
\toprule
\multicolumn{2}{c}{One token (50 entities)} & \multicolumn{2}{c}{Two tokens (50 entities)} \\
\cmidrule(lr){1-2}\cmidrule(lr){3-4}
Einstein & Picasso & Emily Dickinson & Vera Rubin \\
Newton & Gandhi & Mary Shelley & John Locke \\
Darwin & Mandela & Oscar Wilde & Adam Smith \\
Tesla & Lincoln & Victor Hugo & Benjamin Franklin \\
Euler & Churchill & Toni Morrison & George Washington \\
Gauss & Thatcher & Barack Obama & Brad Pitt \\
Turing & Merkel & Rosa Parks & Tom Cruise \\
Kepler & Macron & Thomas Jefferson & Bruce Lee \\
Edison & Netanyahu & Eleanor Roosevelt & Julia Roberts \\
Nobel & Erdogan & Anne Frank & Emma Watson \\
Fibonacci & Reagan & Cristiano Ronaldo & Emma Stone \\
Fourier & Nixon & Michael Jordan & Jennifer Lawrence \\
Aristotle & Truman & Kobe Bryant & Taylor Swift \\
Plato & Eisenhower & Serena Williams & Lady Gaga \\
Kant & Putin & Michael Phelps & Whitney Houston \\
Freud & Stalin & Muhammad Ali & Robert Frost \\
Nietzsche & Lenin & LeBron James & William Blake \\
Shakespeare & Hitler & Marilyn Monroe & John Milton \\
Dickens & Messi & Beyoncé & James Joyce \\
Tolkien & Spielberg & Freddie Mercury & Harper Lee \\
Kafka & Elvis & Robert Boyle & Stephen King \\
Orwell & Madonna & John Dalton & Bob Dylan \\
Trump & Rihanna & Neil Armstrong & Johnny Cash \\
Mozart & Modi & Sally Ride & Louis Armstrong \\
Bach & Oprah & Rachel Carson & Miles Davis \\
\bottomrule
\end{tabular*}
\end{table}

\subsection{Filtering criterion for genie-in-a-bottle}
When applying genie-in-a-bottle, which requires removing context tokens'
attention to entity tokens, we filter out examples for which this removal
produces \texttt{wrong\_name} or \texttt{malformed} outputs before applying
the sliding layer window.